\documentclass[letterpaper, 10 pt, conference]{ieeeconf}  

\IEEEoverridecommandlockouts                              

\usepackage{cite}
\usepackage{amsmath,amssymb,amsfonts}
\usepackage{algorithmic}
\usepackage{graphicx}
\usepackage{textcomp}
\usepackage{xcolor}
\usepackage{url}

\newcommand\numberthis{\addtocounter{equation}{1}\tag{\theequation}}
\title{\LARGE \bf
Technical Report on: Anchoring Sagittal Plane Templates in a Spatial Quadruped
}

\author{Timothy Greco and Daniel E. Koditschek
\thanks{*This work was supported by AFRL grant FA865015D1845 (subcontract 669737-1) and ONR grant \#N00014-16-1-2817, a Vannevar Bush Fellowship held by the second author, sponsored by the Basic Research Office of the Assistant Secretary of Defense for Research and Engineering.}
\thanks{The authors are with the GRASP Lab, University of Pennsylvania, Philadelphia, PA 19014.
        {\tt\small \{tmgreco, kod\} @ seas.upenn.edu}}%
}

\begin{document}
\maketitle
\begin{abstract}
    This technical report provides a more thorough treatment of the proofs and derivations in \cite{Greco_Koditschek_23}. The description of the anchoring controller is reproduced here without abridgement, and additional appendices provide a clearer account of the implementation details.
\end{abstract}
\section{Anchoring Controller}\label{sec:anch}

This anchoring controller is designed to admit interoperable parallel composition with any template whose dynamics renders the sagittal plane invariant by imposing almost global attraction down to that submanifold while avoiding any interference with the particular targeted planar subsystem.  In this paper, the term \textit{template} denotes a second-order dynamical system on the sagittal plane, which is realized by a set of forces $(u_x, u_z)$ applied at a ``virtual toe'' in the sagittal plane.  The anchoring determines a ``virtual toe'' location in the real world, calculates additional forces and moments necessary for stabilizing the robot's out-of-plane degrees of freedom, and returns toe forces that produce the resulting wrench on the robot.

When two paired legs are in contact with the ground, define the \textit{stance frame} as the right-handed inertial coordinate frame with the $z-$axis pointing up and the $y-$axis pointing from the right toe to the left toe, with the origin at the midpoint between the toes. 
Define the \textit{body frame} as the right-handed coordinate frame with the $z-$axis normal to the robot's dorsal plane and the $y-$axis pointing from the robot's right hip to its left, with the origin at the robot's center of mass.  The term {\em pose} will denote the transformation in $SE(3)$ that maps the stance frame to the body frame. 
Define the set $\mathcal{S} \subset SE(3)$ of \emph{sagittal poses} as $\mathcal{S} := \{(p_x, 0, p_z)\} \times \mathcal{P}$, where $\mathcal{P}$ denotes the {\em pitches} --- i.e., the set of rotations around the stance frame that fix the y-axis ---  and define the \emph{sagittal velocities}, $\mathcal{V} \subset \mathbb{R}^6$, as $\mathcal{V} = \{(\dot{p}_x,\,0,\,\dot{p}_z,\,0,\, \alpha,\,0)\}$.  This anchoring controller must make $\mathcal{S} \times \mathcal{V}$ attracting and invariant under the resulting closed loop dynamics. 

Assume that the robot behaves as a single rigid body with massless legs, and that two feet are in contact with the ground, either the front feet or the hind feet.  Suppose further that the robot's legs do not slip, reach kinematic singularity, or require more torque than the motors can provide.  In this configuration, the robot can directly actuate five of the torso's six degrees of freedom.  It cannot directly produce a torque about the line between the toes, i.e. in the pitch direction, so this rotational degree of freedom must remain coupled to the $x$ and $z$ translational degrees of freedom.  Templates defined in the sagittal plane expect and account for this coupling between $x$, $z$, and pitch, so this anchoring controller reserves those degrees of actuation for the template controller.  Since the remaining degrees of freedom are fully actuated, the anchoring can use one controller to stabilize the lateral position while a second controller stabilizes the orientation.

Let the lateral translation controller take the familiar potential-dissipative form (proportional-derivative, for these translational components of the pose), 
\begin{equation}
\label{eqn:lat}
    u_y = -K_p p_y - K_d \dot{p_y},
\end{equation}
where $u_y$ is a lateral force applied on the robot at the midpoint of the toes and $p_y$ is the body position in the $y-$direction of the stance frame.  If the robot can maintain contact with the ground and if gravity has a negligible effect on the robot's lateral movement, this controller must stabilize the robot's position to the sagittal plane.  These assumptions match the observed behavior documented in \cite{Greco_Koditschek_23}.  Thus, it remains to introduce a controller stabilizing the pitches, $\mathcal{P}$, under the dynamical system on $TSO(3)$:
\begin{align*}
    \dot{R} &= RJ(\omega)^T \\
    \dot{\omega} &=  M^{-1}(\tau - \omega \times M\omega), \numberthis \label{eqn:dynsys}.
\end{align*}
This system can also be stabilized by a potential-dissipative controller \cite{KOD89} of the form 
\begin{equation}
\label{eqn:rot}
    \tau = -\nabla \Phi(R) - K_D\omega, 
\end{equation}
where the rotation matrix $R$ represents the angular component of the pose, $\omega \in \mathbb{R}^3$ is its angular velocity, and the controller terms are parametrized as 
\begin{equation} 
\label{eqn:phidef}
   K_D = \begin{bmatrix}\kappa_1 & 0 & 0 \\ 0 & 0 & 0 \\ 0 & 0 & \kappa_2 \end{bmatrix}\quad\text{ and }\quad \Phi(R) = y^TRy
\end{equation}
for $y = [0,\ 1,\ 0]^T$ and $\kappa_1, \kappa_2 > 0$.  The following subsections formally demonstrate that the controller defined in (\ref{eqn:rot}) makes $T\mathcal{P} \subset TSO(3)$ attracting and invariant. 

This demonstration is organized as follows.  In Section \ref{sec:grad}, calculating the gradient of $\Phi$ confirms that $\mathcal{P}$ lies in its critical set, so $T\mathcal{P}$ is an equilibrium of the closed-loop dynamics (\ref{eqn:dynsys}) arising from (\ref{eqn:rot}) and thus it is invariant.  Section \ref{sec:hess} continues with the calculation of the Hessian of $\Phi$ at the components of its critical set.  Section \ref{sec:lasalle} uses total energy as a Lasalle function to show that the critical set of $\Phi$ is globally attracting.  Section \ref{sec:local} concludes with an examination of the local stability of equilibrium points in each component of the critical set, applying the results from Section \ref{sec:hess} to conclude that only point equilibria in $T\mathcal{P}$ are locally attracting, while point equilibria elsewhere in the critical set are unstable. 
The results of Sections \ref{sec:lasalle} and \ref{sec:local} establish that $T\mathcal{P}$ is “almost globally” attracting in the sense of \cite{KOD89} when the only invariant set generated by the template dynamics in the sagittal plane consists of local equilibrium states.  Some of the sagittal template dynamics we use in this work have this property,  whereby the combination of global (\ref{sec:lasalle}) and local (\ref{sec:local}) results suffice to guarantee that only a zero measure set of initial conditions generate trajectories that fail to converge to the desired sagittal plane behavior in $T\mathcal{P}$.  We conjecture (but do not prove in this paper) that this controller ensures almost global convergence to $T\mathcal{P}$ regardless of the properties of the template dynamics.  If this conjecture is sound, as our empirical results suggest, then it formally guarantees global efficacy  of this anchoring controller in parallel composition with any desired sagittal-plane behavior (including, for example,  the steady state bounding template that introduces a non-equilibrium attracting invariant set, i.e. a hybrid limit cycle) on any quadruped with twelve or more actuated degrees of freedom.
\subsection{The Gradient of $\Phi$}\label{sec:grad}
Consider the gradient $\nabla_R \Phi$ with respect to $\omega \in \mathbb{R}^3$.  Apply equation (3.1) from \cite{bozma_19}, recalling that $y^TRy = \mathbf{Tr}(yy^TR)$:
\begin{equation}
\label{eqn:graddef}
    \nabla_R \Phi (\omega) = \frac{1}{2}\mathbf{Tr}\big((yy^T - R^Tyy^TR^T)RJ(\omega)\big),
\end{equation}
where $J: \mathbb{R}^3 \to \mathfrak{so}(3)$ is the skew map such that $$J(a)b = a \times b.$$  After some massaging,
\begin{align*}
    \nabla_R \Phi (\omega) &= \frac{1}{2}\mathbf{Tr}\big((yy^T - R^Tyy^TR^T)RJ(\omega)\big) \\
    &= \frac{1}{2}\mathbf{Tr}\big(yy^TRJ(\omega) - R^Tyy^TJ(\omega)\big) \\
    &= \frac{1}{2}\mathbf{Tr}\big(y^TRJ(\omega)y - y^TJ(\omega)R^Ty\big) \\
    &= \frac{1}{2}\big(y^TR(\omega \times y) - y^T(\omega \times R^Ty)\big) \\
    &= \frac{1}{2}\big(\omega^T(y \times R^Ty) - \omega^T(R^Ty \times y)\big) \\
    &= \omega^T(y \times R^Ty)
\end{align*}
yields an expression of $\nabla_R \Phi\in \mathbb{R}^3$ as
\begin{equation}
\label{eqn:grad}
    \nabla_R \Phi = y \times R^Ty.
\end{equation}

\begin{figure}[t!]
    \centering
    \includegraphics[width=0.49\textwidth]{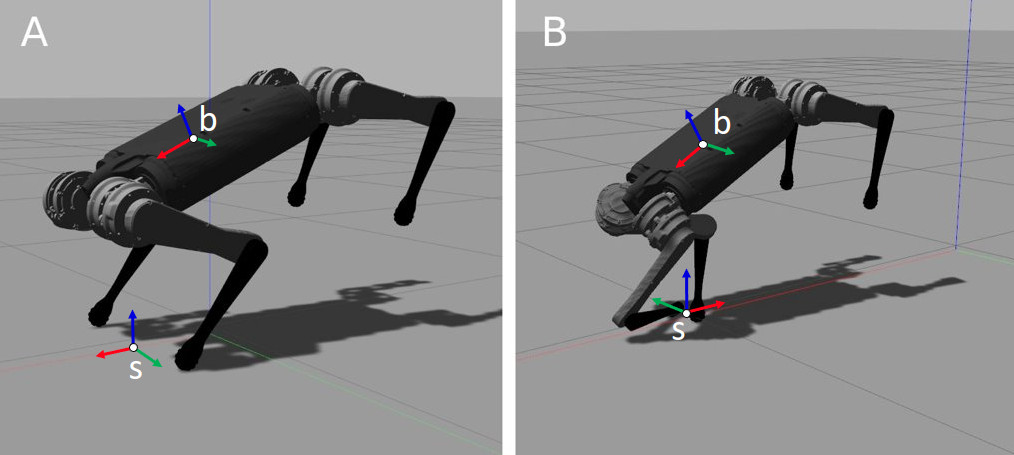}
    \caption{Visualizing the critical set of $\Phi$.  A displays a pose is in $\mathcal{P}$, and B displays a pose in $\mathcal{Q}$.  This latter configuration is kinematically feasible but difficult to achieve.  In each picture, the stance frame is labeled `s' and the body frame `b'.}
    \label{fig:PQ}
\end{figure}

This formulation makes it easy to determine the critical regions of $\Phi$, since $\nabla_R \Phi = 0$ if and only if $R^Ty = \lambda y$ for $\lambda \in \{-1, 1\}$.  Now $R^Ty = y$ if and only if $R \in \mathcal{P}$; hence, it is useful to define the corresponding antipodal set as $\mathcal{Q}:= \{R\in SO(3)\ |\ R^Ty = -y\}$.  
These two disjoint sets comprise the critical set of $\Phi$.  Since for any $(R,\omega) \in T\mathcal{P}$, $R$ is in the critical set of $\Phi$ and $\omega$ is in the null space of $K_D$, so the controller defined in (\ref{eqn:rot}) exerts zero input, and thus $T\mathcal{P}$ is invariant under this controller.  Fig. \ref{fig:PQ} shows what configurations in $\mathcal{P}$ and $\mathcal{Q}$ actually look like for the robot.

\subsection{The Hessian of $\Phi$}\label{sec:hess}
Lemma 3.2 from \cite{bozma_19} gives an expression for the Hessian of $\Phi$ at any critical point as a symmetric bilinear form on $\mathfrak{so}(n)$.  Applying this result to (\ref{eqn:phidef}) yields
\begin{equation}
    h_\Phi(U,V) = -\mathbf{Tr}(yy^TRUV)
\end{equation}where $U,V \in \mathfrak{so}(3)$.  To determine the relative maxima and minima of $\Phi$, evaluate the Hessian for some $U = V$.  Since $\mathfrak{so}(3)$ is isomorphic to $\mathbb{R}^3$, there is some $u$ such that $J(u) = U$.  This isomorphism enables the representation of the Hessian as a quadratic form on $\mathbb{R}^3$, simplifying its evaluation.
\begin{align*}
h_\Phi(u,u) &= -\mathbf{Tr}(yy^TRJ(u)J(u) \\
    &= -y^TRJ(u)J(u)y\\
     &= (R^Ty)\cdot \big(u \times J(u)y\big) \\
     &= -\big(J(u)y\big) \cdot \big( R^Ty \times u \big) \\
     &= \big(J(y)u\big) \cdot J(R^Ty) u \\
     &= u^TJ(y)^TR^TJ(y)Ru \\
     &= u^TJ(y)^TJ(R^Ty)u
\end{align*}
The resulting Hessian matrix is 
\begin{equation} 
\label{eqn:hess}
    H_\Phi(R) := J(y)^TJ(R^Ty).
\end{equation}
If $R$ is in the critical set, $H_\Phi(R) = \lambda J(y)^TJ(y)$.  
Evaluating this equation for $R \in \mathcal{P}$,
\begin{equation}
    H_\Phi(R)  = \begin{bmatrix}1 & 0 & 0 \\ 0 & 0 & 0 \\ 0 & 0 & 1 \end{bmatrix} =: H_\mathcal{P}.
\end{equation}
Similarly, for $R \in \mathcal{Q}$,
\begin{equation}
    H_\Phi(R) = \begin{bmatrix}-1 & 0 & 0 \\ 0 & 0 & 0 \\ 0 & 0 & -1 \end{bmatrix} =: H_\mathcal{Q}.
\end{equation}

Thus $\mathcal{P}$ is a minimum and $\mathcal{Q}$ a maximum of $\Phi$ on $SO(3)$, and since $SO(3)$ has no boundary, these extrema are global.  The vector $y$ spans both the kernel of these Hessian matrices and the tangent spaces of both $\mathcal{P}$ and $\mathcal{Q}$, and thus $\Phi$ is a Morse-Bott function on $SO(3)$ \cite{Bott}.
\subsection{$\Phi$ as a potential}\label{sec:lasalle}
The dynamical system defined in (\ref{eqn:dynsys}), with $\tau$ is defined as in (\ref{eqn:rot}), describes the effect of the orientation controller on the robot in the absence of any other torques.  
Define the stance-frame moment of inertia $M(R) = R^T I_B R$, where $I_B$ denotes the diagonal matrix that expresses the robot's inertia tensor in the body frame.  Consider the total energy function $\eta:\ TSO(3) \to \mathbb{R}$ defined as
\begin{equation}
\label{eqn:LSfun}
    \eta(R,\omega) = \Phi(R) + \frac{1}{2}\omega^T M(R) \omega,
\end{equation}
where $\omega$ is the robot's angular velocity in the stance frame.  This function is a Lasalle function for the system defined in (\ref{eqn:dynsys}) \cite{LaSalle}, \cite{khalil2002nonlinear}.  To see this, first note that $\eta(R,\omega) \geq 0$, since $\Phi(R) \geq 0$ and $M$ is symmetric positive definite.  
Examining $\dot{\eta}(R,\omega)$,
\begin{equation}
\label{eqn:LSfundot}
    \dot{\eta}(R,\omega) = \omega^T(y \times R^Ty) + \omega^TM \dot{\omega} + \frac{1}{2}\omega^T\dot{M}\omega.
\end{equation}
Since $M = R^T I_B R$,
\begin{equation}
\label{eqn:Mdot}
    \omega^T \dot{M} \omega = 2\omega^T \dot{R^T} I_B R \omega = 2\omega^T (\omega \times M \omega) = 0.
\end{equation}
Substituting (\ref{eqn:rot}), (\ref{eqn:Mdot}), and (\ref{eqn:dynsys}) into (\ref{eqn:LSfundot}),
\begin{align*}
    \dot{\eta}(R,\omega) &= \omega^T(y \times R^Ty) \\
    &\quad+ \omega^TMM^{-1}( -\nabla \Phi(R) - K_D\omega - \omega \times M\omega) \\
    &= \omega^T(y \times R^Ty) - \omega^T \nabla\Phi(R) - \omega^T K_D \omega \\
    &\quad - \omega^T( \omega \times M\omega)\\
    &= \omega^T(y \times R^Ty) - \omega^T \nabla\Phi(R) - \omega^T K_D \omega.
\end{align*}
Applying (\ref{eqn:grad}) yields
\begin{equation}
\label{eqn:LSfundotresult}
    \dot{\eta}(R,\omega) = - \omega^T K_D \omega,
\end{equation}
and since $K_D$ is positive semidefinite, $\dot{\eta} \leq 0$. Note that $\dot{\eta} = 0$ if and only if $\omega \in \mathrm{span}(y)$ (i.e., $\omega = \alpha y$ for some $\alpha \in \mathbb{R}$) in the stance frame. 

By Lasalle's invariance principle, the largest invariant subset of $\eta^{-1}[0]$ is attracting and contains all forward limit points of this dynamical system; it remains to be shown whether this subset contains $T\mathcal{P}$.  Let $\mathcal{L} \subset \eta^{-1}[0]$ be this largest invariant subset.  For any $(R, \omega) \in \mathcal{L}$, the angular acceleration must be in the $y$ direction to remain in that set under the flow of \ref{eqn:dynsys}.  If $\omega = \alpha y$ for some $\alpha \in \mathbb{R}$, then
\begin{align*}
    \dot{\alpha}y &= \dot{\omega} \\
    &= M^{-1}\!\big(\!-\!\nabla\Phi(R) - K_D(\alpha y) - (\alpha y) \times M(\alpha y)\big) \\
    \dot{\alpha}My &= -(y \times R^Ty) - \alpha^2y \times My. \numberthis \label{eqn:zeroacc}
\end{align*}
Since $M$ is symmetric positive definite, $My$ must have a nonzero component in the $y$ direction.  However, neither of the terms on the RHS of (\ref{eqn:zeroacc}) can have any nonzero component in the $y$ direction, so (\ref{eqn:zeroacc}) can only hold if $\dot{\alpha} = 0$, from which it follows that 
\begin{equation}
\label{eqn:zeroacc2}
    0 = -(y \times R^Ty) - \alpha^2y \times My.
\end{equation}
To completely characterize the rotational trajectories $\big(R(t),\,\omega(t)\big)$ of the flow generated by (\ref{eqn:dynsys}) that satisfy (\ref{eqn:zeroacc2}), it is helpful to impose the consequent necessary requirement that time variation of (\ref{eqn:zeroacc2}) also evaluate to the constant 0.  Accordingly, taking the derivative of (\ref{eqn:zeroacc2}) with respect to time yields
\begin{align*}
    0 &= -y \times R(\alpha y \times y) - 2\alpha\dot{\alpha}y \times My - \alpha^2\big(y \times (\alpha y \times My)\\
    &\quad\;\! + y \times M(\alpha y \times y)\big) \\
    0 &= -\alpha^3\big(y \times (y \times My)\big). \numberthis \label{eqn:derivzero}
\end{align*}
Since $y$ must be perpendicular to $y \times My$ if the latter is nonzero, (\ref{eqn:derivzero}) can only be satisfied if $\alpha = 0$ or $y \times My = 0$.  In either case, (\ref{eqn:zeroacc2}) simplifies to \begin{equation}
    0 = -y \times R^Ty = -\nabla\Phi(R). \\
\end{equation}
Thus $(R, \omega) \in \mathcal{L}$ if and only if $R \in \mathcal{P} \cup \mathcal{Q}$ and $\omega \in \mathrm{span}(y)$.  The embeddings in $\mathbb{R}^3$ of the tangent spaces of both $\mathcal{P}$ and $\mathcal{Q}$ are both represented by $\mathrm{span}(y)$ in the stance frame, so $\mathcal{L} = T\mathcal{P} \cup T\mathcal{Q}$.  By Lasalle's invariance principle \cite{LaSalle}, \cite{khalil2002nonlinear}, $T\mathcal{P} \cup T\mathcal{Q}$ is attracting and contains all forward limit points.

\subsection{Local Stability of Equilibria}\label{sec:local}

Suppose that, in addition to the anchorning controller (\ref{eqn:rot}), the robot is subject to some template dynamics on $T\mathcal{P}$; further suppose that the template is ``pitch-steady'' as described in \cite{de_topping_caporale_koditschek_22}, so there will be some $p_0 \in \mathcal{P}$ that is locally attracting within $T\mathcal{P}$.  The template dynamics around $p_0$ in $T\mathcal{P}$ can be approximated by a second-order linear system 
\begin{equation}
    \label{eqn:lintemp}
    \begin{bmatrix} \dot{p} \\ \ddot{p} \end{bmatrix} = \begin{bmatrix}
    0 & 1 \\ -\frac{\gamma}{\mu} & -\frac{\beta}{\mu}
    \end{bmatrix}\begin{bmatrix}
    p \\ \dot{p}
    \end{bmatrix},
\end{equation}
where $\beta$, $\gamma$, and $\mu$ are positive constants.  Since $T\mathcal{P}$ is invariant under (\ref{eqn:rot}), the anchoring will not alter this behavior.  Accordingly, the linearized dynamics about $p_0$ in $TSO(3)$ are
\begin{equation}
    \label{eqn:linear_p}
    \begin{bmatrix} \dot{\theta} \\ \dot{\omega} \end{bmatrix}\! = \!\begin{bmatrix}
    0 & I \\ -M^{-1}(H_\mathcal{P} + {K})  & -M^{-1}(K_D + {B})
    \end{bmatrix}\!\!\begin{bmatrix}
    \theta \\ \omega
    \end{bmatrix},
\end{equation}
where $K = \mathrm{diag}([0, \gamma, 0])$ and $B = \mathrm{diag}([0, \beta, 0])$.  Since $K_D + B$ and $M$ are symmetric positive definite, and $H_\mathcal{P} + K$ is positive definite, (\ref{eqn:linear_p}) is asymptotically stable \cite[Lemma 3.5]{koditschek_89}.  Thus $p_0$ is an attractor.

Conversely, consider the action of the same template dynamics when the robot's pose and velocity lie in $T\mathcal{Q}$.  There will be some corresponding equilibrium point $q_0 \in \mathcal{Q}$ which admits the same linearized dynamics as in (\ref{eqn:lintemp}).  In this case, the linearized dynamics about $q_0$ on $TSO(3)$ are
\begin{equation}
    \label{eqn:linear_q}
    \begin{bmatrix} \dot{\theta} \\ \dot{\omega} \end{bmatrix}\! = \! \begin{bmatrix}
    0 & I \\ -M^{-1}(H_\mathcal{Q} + {K})  & -M^{-1}(K_D + {B})
    \end{bmatrix}\!\!\begin{bmatrix}
    \theta \\ \omega
    \end{bmatrix}.
\end{equation}
In this system, $H_\mathcal{Q} + {K}$ has both positive and negative eigenvalues so $q_0$ is a saddle \cite[Lemma 3.5]{koditschek_89}.  

These local stability results clarify the results of the previous section.  The equilibrium point of any pitch-stable template is stable in $\mathcal{P}$ but unstable in $\mathcal{Q}$, suggesting that of the two disjoint sets that compose a global attractor on $SO(3)$, only $\mathcal{P}$ is locally attracting.  While the extension of this result to limit cycles or abitrary trajectories in pitch remains unproven, the empirical results of \cite{Greco_Koditschek_23} support the plausibility of such a conjecture.

\section{Further Implementation Details}\label{app:explan}
Assuming a single rigid body model of the robot's torso, let $(x,y,z)$ be the translational degrees of freedom in the stance frame and let $(\theta, \phi, \psi)$ be the roll, pitch, and yaw.  The template controllers specify forces $u_x$ and $u_z$ to be exerted at a virtual toe, and thus control the robot's motion in the $x$, $z$, and $\phi$ directions.  The anchoring controllers specify a lateral force $u_y$ and a torque $\tau$, and aim to stabilize the robot's motion in $y$, $\theta$, and $\psi$.  This appendix explains how the implementation maps from these controller outputs to toe forces, which are represented in sum and difference coordinates.  Here $r = [x, y, z]^T$ is the vector from the COM to the midpoint between the toes and $k$ is the vector from the midpoint between the toes to the left toe.  In the stance frame, $k$ points directly in the $y$ direction.
Assume that the robot has two feet are in contact with the ground, either the front feet or the hind feet, and that each leg can exert an arbitrary force in $\mathbb{R}^3$ on the body without slipping or lifting off the ground.  Let $f_l$ and $f_r$ be the forces at the left and right toes respectively, and let $x_l$ and $x_r$ be their locations with respect to the center of mass.  Ignoring gravity and coriolis terms, Newton's second law yields
\begin{align*}
    m\ddot{r} &= f_l + f_r \\
    M\dot{\omega} &= x_l \times f_l + x_r \times f_r,
\end{align*}
where $m$ is the mass and $M$ is the moment of inertia matrix as represented in the stance frame. Translating this into sum and difference coordinates will make it easier to separate the forces acting on the center of mass from the torques.  Let $s = f_l + f_r$ and $d = f_l - f_r$.  It follows from the definition of the stance frame that $r = \frac{1}{2}(x_l + x_r)$, and that $k =  \frac{1}{2}(x_l - x_r)$ points in the direction of the $y$ axis.  Inverting these transformations yields $f_l = \frac{1}{2}(s + d)$ and $f{_r} = \frac{1}{2}(s - d)$.  Similarly, $x_l = r + k$ and $x_r = r - k$.  Thus
\begin{align*}
    m\ddot{r} =& s \\
    M\dot{\omega} =& (r + k) \times \frac{1}{2}(s + d) + (r - k) \times  \frac{1}{2}(s - d)\\
    =& \frac{1}{2}\Big(r \times s + k \times s + r \times d + k \times d + r \times s \\
    &{-}\:k \times s - r \times d + k \times d \Big)\\
    =&\frac{1}{2}\Big(2\big(r \times s\big) + 2\big(k \times d)\Big) \\
    =&r \times s + k \times d.
\end{align*}
 Assume that the robot's orientation is close enough to the sagittal plane that the matrix $M$ is approximately diagonal.  Since $k$ is in the $y$ direction, $(k \times d)$ has no $y$ component and thus the only acceleration in the pitch direction will be from $r \times s$.  Similarly, the $y$ component $s_y$ will not affect the pitch dynamics.  Thus set $s_x = u_x$ and $s_z = u_z$ to execute the template dynamics in the sagittal-plane degrees of freedom $x, z, $ and $\phi$.

Similarly, $s_y$ is the only input that exerts a force the $y$ direction, so set $s_y = u_y$.  While $\theta$ and $\psi$ are affected by both $d$ and $s$, in practice the torque induced by $r \times s$ is almost entirely in the pitch direction.

Let $r = r_p + r_y$, where $r_p$ lies in the $xz$ plane and $r_y$ points in the $y$ direction. Similarly, let $s_{xz}$ be the sagittal plane component of $s$ and let $s_y$ be the $y$-component.  Then
\begin{align*}
    r \times s &= r_p \times s + r_y \times s \\
    &= r_p \times s_{xz} + r_y \times s_{xz} + r_p \times s_y + r_y \times s_y
\end{align*}
The vectors $r_y$ and $s_y$ are parallel, so their cross product is zero.  The terms $ r_y \times s_{xz} + r_p \times s_y $ are both perpendicular to the $y$ axis, so they can be directly cancelled out by the $k \times d$ term.  In practice, $r_y$ and $s_y$ are small enough that the effect of this term can be neglected, admitting the approximation
$$ s \times r \approx r_p \times s_{xz}$$
as only effecting the pitch.  This leaves the $k \times d$ as the only input that controls the other rotational degrees of freedom.  Thus the torque exerted by the anchoring controller is
$$ \tau = \begin{bmatrix}
    \tau_\theta \\ 0 \\ \tau_\psi
\end{bmatrix} = k \times d,$$
so $d = J(k)^\dagger\tau$ yields a value for $d$ that will generate the desired body torques, where $J(k)$ is a skew-symmetric matrix and $J(k)^\dagger$ denotes its pseudoinverse.  The $y$ component of $d$ is set to zero to eliminate internal forces between the toes.
\section{Assumptions}
\begin{enumerate}
    \item The dynamics are that of a single rigid body; the legs have negligible mass.  Furthermore, the legs are assumed to not collide with each other and to not reach singularity.
    \item The robot has two feet in contact with the ground, either the forelegs or the hind legs.  When all four legs are in contact with the ground, there may not be consensus between the two halves on which plane is the sagittal plane, causing internal forces and hindering the robot from converging to either.  However, these effects were not noticeable during double-stance periods, such as those occurring during the pronk or box jump.
    \item Each of the two toes in contact with the ground is fixed in the world frame and can exert an arbitrary force in $\mathbb{R}^3$ on the robot's body.
    \begin{itemize}
        \item This assumption is violated when the robot's toes slip.  For the most part, the only times when slipping occurs are when the toes bounce off the ground during the steady-state gaits or when the robot is recovering from large disturbances in yaw.
        \item The assumption of arbitrary force in $\mathbb{R}^3$ guarantees that the robot can exert whatever force is necessary for the anchoring controller.  Blindly following this assumption could cause the robot to pull one of its toes off the ground to respond to a disturbance in roll.  The code does take this into account, and will scale down the input from the anchoring controller to prevent it.
    \end{itemize}
    \item Gravity has a negligible effect outside of the sagittal plane, i.e. the ground is flat and the center of mass is close to the sagittal plane (so there is little action of gravity in the roll and $y$ directions).
    \item The stability analysis assumes that all of these assumptions are true for an infinite amount of time, so will be violated by cases when, for example, the trajectory of the robot makes it kinematically unable to maintain contact with the ground (such as very high lateral velocity).
    \item As written above, $s \times r$ has negligible components outside of the pitch direction.
    This assumption is not necessary, but it simplifies the implementation.
    
\end{enumerate}

\bibliography{refs2}{}
\bibliographystyle{ieeetr}
\end{document}